\documentclass[letterpaper, 10 pt, conference]{ieeeconf}  

\IEEEoverridecommandlockouts                              

\usepackage{cite}
\usepackage{graphicx}
\usepackage{subcaption}
\usepackage{multirow}
\usepackage{color}

\title{\LARGE \bf
DerainNeRF: 3D Scene Estimation with Adhesive Waterdrop Removal
}

\author{Yunhao Li$^{1,2}$, Jing Wu$^{2}$, Lingzhe Zhao$^{2}$ and Peidong Liu$^{2*}$
\thanks{This work was supported in part by NSFC under Grant 62202389, in part by a grant from the Westlake University-Muyuan Joint Research Institute, and in part by the Westlake Education Foundation.}
\thanks{*Corresponding author}
\thanks{$^{1}$Yunhao Li is with School of Computer Science and Engineering, Zhejiang University, No. 866 Yuhangtang Road, Hangzhou, Zhejiang, China. {\tt\small yunhaoli@zju.edu.cn}} 
\thanks{$^{2}$Yunhao Li, Jing Wu, Lingzhe Zhao and Peidong Liu are with School of Engineering, Westlake University, No. 600 Dunyu Road, Hangzhou, Zhejiang, China. {\tt\small \{wujing05, zhaolingzhe, liupeidong\}@westlake.edu.cn}
        }%
}

\begin{document}

\maketitle
\thispagestyle{empty}
\pagestyle{empty}

\begin{abstract}

When capturing images through the glass during rainy or snowy weather conditions, the resulting images often contain waterdrops adhered on the glass surface, and these waterdrops significantly degrade the image quality and performance of many computer vision algorithms. To tackle these limitations, we propose a method to reconstruct the clear 3D scene implicitly from multi-view images degraded by waterdrops. Our method exploits an attention network to predict the location of waterdrops and then train a Neural Radiance Fields to recover the 3D scene implicitly.
By leveraging the strong scene representation capabilities of NeRF, our method can render high-quality novel-view images with waterdrops removed. Extensive experimental results on both synthetic and real datasets show that our method is able to generate clear 3D scenes and outperforms existing state-of-the-art (SOTA) image adhesive waterdrop removal methods.

\end{abstract}

\section{INTRODUCTION}

\begin{figure}[h]
  \centering
  
    \includegraphics[width=0.99\linewidth]{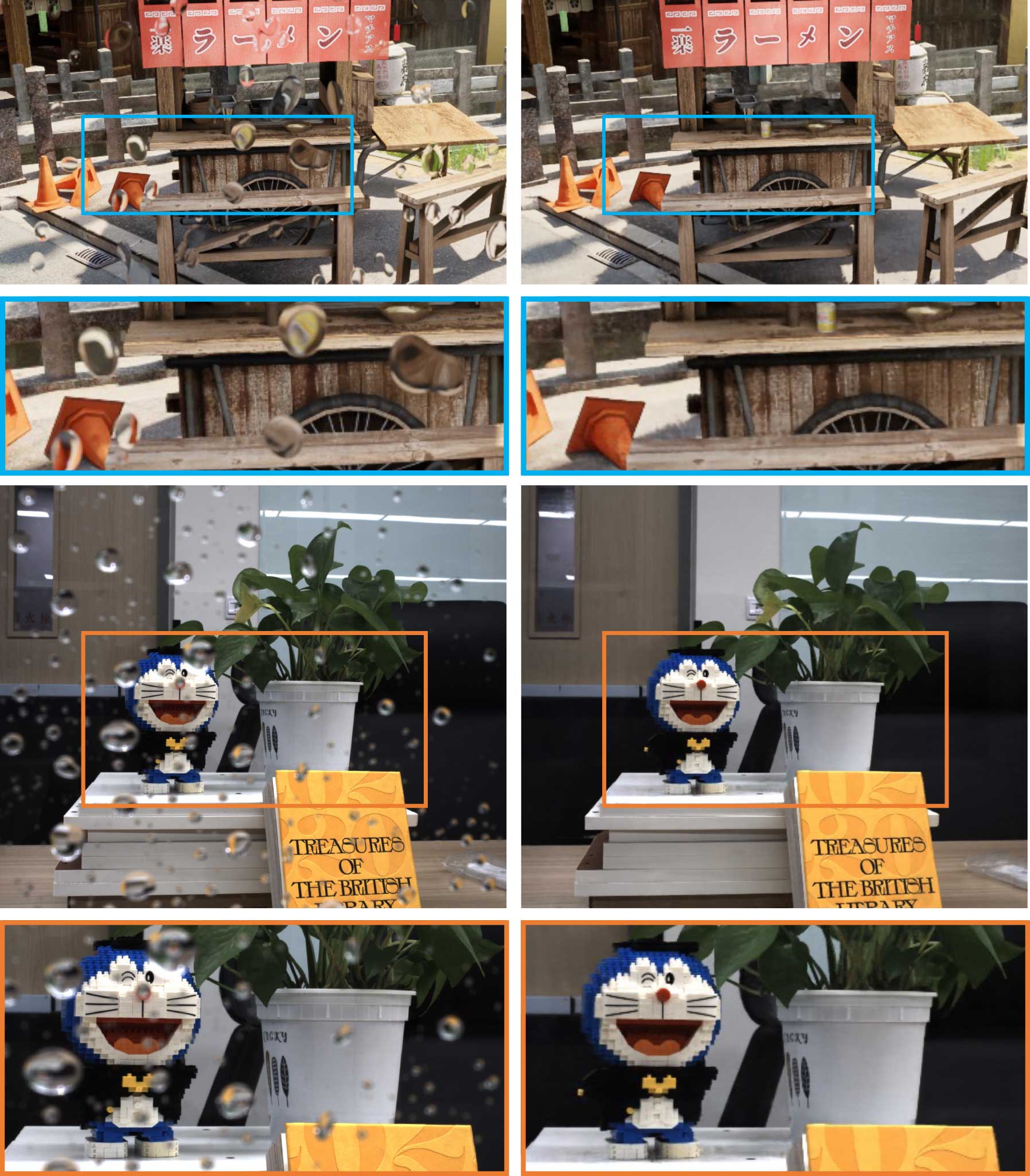}
    \caption{Given a set of waterdrop images (left column), our DerainNeRF estimates 3D scenes and reomves the adhesive waterdrops altogether. It synthesizes clear images (right column) with high quality.}
    \label{fig1}
    \vspace{-1em}
\end{figure}

3D scene representation and estimation techniques have wide range of applications in autonomous driving, robotics, virtual reality (VR) and cultural heritage preservation. In recent years, Neural Radiance Fields (NeRF)\cite{mildenhall2021nerf} have gained popularity for 3D reconstruction and scene representation due to their ability to provide continuous scene representation, robustness in handling complex scenes, and state-of-the-art performance on novel-view image synthesis. However, in many real-world scenarios, particularly those involving outdoor images like autonomous driving and drones, it is likely that the images taken under rainy or snowy weather conditions come with adhesive raindrops, as illustrated in the top row of Fig. \ref{fig1}. Those images will further degrade the performance of related applications, such as 3D reconstruction \cite{mildenhall2021nerf}, visual perception\cite{bruls2018mark}\cite{valada2017adapnet}, object detection\cite{wang2020overview}\cite{mao20223d}, and tracking\cite{qian20223d}\cite{arnold2019survey}. Therefore, to tackle those limitations, we propose to simultaneously remove the adhesive waterdrops from captured images and recover the underlying 3D scene implicitly in this work, by leveraging the impressive 3D scene representation capability of NeRF.

The vanilla NeRF framework is not robust to images with adhesive waterdrops. Although many researchers have proposed dedicated methods to make NeRF robust under in-the-wild scenarios (e.g. challenging illumination conditions\cite{martin2021nerf}, dynamic environments\cite{park2021hypernerf}), they cannot handle the images with adhesive waterdrops well because of the random spatial distribution, irregular shapes, and complicated refraction and reflection properties of waterdrops. 
To address the issue of waterdrop removal, we propose a NeRF-based framework that simultaneously estimates the 3D scenes while removing waterdrops. We refer this model as DerainNeRF, i.e., NeRF with waterdrop removal function. 

Inspired by existing waterdrop removal methods, and dedicated NeRF frameworks that handle image occlusion\cite{zhu2023occlusion} and scene object removal\cite{weder2023removing}\cite{wei2023clutter}, DerainNeRF combines a waterdrop detection network and NeRF for 3D scene representation learning. In particular, it first exploits a pre-trained deep waterdrop detector to predict the locations of waterdrops. It then excludes the waterdrop covered pixels during the training of NeRF, so that it can recover the clear scenes from non-occluded pixels. 
We evaluate DerainNeRF using both synthetic and real datasets. The experimental results demonstrate that DerainNeRF effectively estimates clear 3D scenes from waterdrop images, and renders novel-view clear images as is shown in Fig. \ref{fig1}. Both the quantitative and qualitative results demonstrate our method delivers superior quality compared to the existing state-of-the-art image waterdrop removal methods. To the best of our knowledge, DerainNeRF is the first NeRF-based method which takes waterdrop degraded images as input and recovers the clear scene implicitly. 


\section{RELATED WORK}
We review two main areas of the prior works: Neural Radiance Fields (NeRF) and image adhesive waterdrop removal, which are the most related to our work.

\subsection{NeRF}
Mildenhall et al.\cite{mildenhall2021nerf} propose NeRF, an epochal 3D scene representation and novel-view image rendering method. Different from discrete voxel-based methods\cite{lombardi2019neural}, NeRF estimates a continuous scene volumetric function via a Multi Layer Perceptron (MLP). With the help of differentiable volumetric rendering technique, NeRF demonstrates impressive novel view image synthesis performance and 3D scene representation capability. To make NeRF robust to more complicated real-world environments, researchers have proposed many dedicated NeRF variants recently. Some researches extend NeRF to make it compatible for large scale scene reconstruction and representation, making it available to autonomous driving and robotics related applications\cite{tancik2022block}\cite{xiangli2023bungeenerf}\cite{turki2022mega}. Others focus on developing NeRF which can deal with inaccurate camera poses and challenging imaging conditions\cite{jeong2021self}\cite{lin2021barf}\cite{wang2021nerf}\cite{ma2022deblur}\cite{wang2023bad}. There are also many other NeRF variants for High Dynamic Range (HDR) image modeling\cite{huang2022hdr} and NeRF for scene editing\cite{yuan2022nerf}\cite{weder2023removing}\cite{wei2023clutter}. 

Pan et al.\cite{pan2022sampling} develop a specialized NeRF-based framework which deals with scene of bounded volume and boundary from multi-view posed images with refractive object silhouettes by extending sampling techniques to drawing samples along a curved path modeled by Eikonal equation\cite{bruss1982eikonal}. WaterNeRF\cite{sethuraman2022waternerf} introduced by Sethuraman et al. deals with underwater scene reconstruction by estimating parameters of a physics-based model for image formation. Wang et al.\cite{wang2022neref} propose NeReF, which estimates the refractive fluid surface with implicit representation. 

For occlusion removal and scene editing, Weder et al.\cite{weder2023removing} propose a NeRF-based scene object removal scheme. Given a user-generated mask, the proposed method first blocks the removed object in input images, impaints the blocked regions using a pre-trained network, then a confidence-driven view-selection scheme enforces multi-view consistencies from the inpainted images. The closest work to ours is Zhu et al.\cite{zhu2023occlusion}, which propose a NeRF-based method with occlusion removal. However, different from focusing on scenarios where occluders are fixed to the scene and camera takes multi-view images, our work can deal with not only the case where the adhesive waterdrops are fixed to the scenes, but also the case when the waterdrops are fixed to the camera. Additionally, our work utilizes a pre-trained rain-detection mechanism from state-of-the-art image waterdrop removal methods, which is significantly simplier than that of Zhu et al. \cite{zhu2023occlusion}, which requires additional scene MLP and depth-based mask MLP to separate clear background.

\subsection{Image Adhesive Waterdrop removal}

Many existing deraining methods focus on removing rain streaks, which has simplier image formation models. However, the physical properties of adhesive waterdrops differ from rain streaks. Earliest methods focus on modeling the waterdrops by estimating its geometric shape, refraction and reflection properties\cite{garg2003photometric}\cite{yu1999new}. Others leverages temporal features\cite{you2015adherent}, for example, optical flow, or spatial features\cite{yamashita2009noises}\cite{tanaka2006removal}, for example, disparities, to separate the waterdrops from the clear background. In recent years, some waterdrop removal methods have been introduced. Eigen et al.\cite{eigen2013restoring} proposes the first end-to-end image adhesive waterdrop removal method based on deep CNN. Due to its relatively simple and shallow network architecture, its performance becomes poor when the area of adhesive waterdrops in the images becomes large. Qian et al.\cite{qian2018attentive} introduce AttGAN, a waterdrop removal model based on generative adversarial network (GAN)\cite{goodfellow2014generative}. In AttGAN, the generator contains an attention module, which generates an attention map, then removes the waterdrops based on both input image and attention map. However, the attention-based mechanism in AttGAN fails to leverage global spatial information. Quan et al.\cite{quan2019deep} develop an approach with shape-driven and channel attention modules, which performs better on large waterdrop removal, but it is still limited to local attention aggregation.  Wen et al.\cite{wen2023video} propose a video/multi-image waterdrop removal methods for complex driving scenes based on spatial-temporal information fusion by self-attention mechanism and a cross-modality training strategy.


\section{METHOD}

\begin{figure*}[!ht]
  \centering
  
    \includegraphics[width=0.98\linewidth]{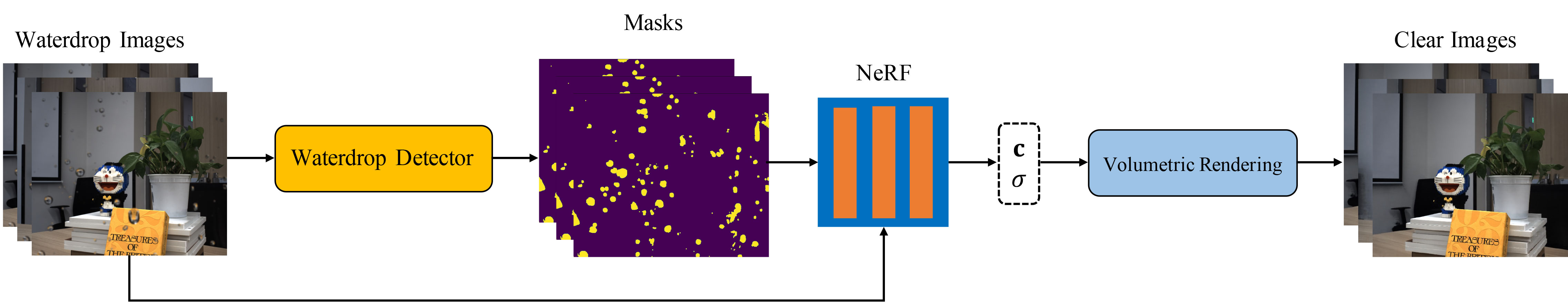}
    \caption{Training procedure of DerainNeRF. A pre-trained deep waterdrop detector detects waterdrops in input images and generate binary masks, then DerainNeRF utilizes the masks to block waterdrop regions in input images during NeRF training.}
    \label{fig2}
    \vspace{-1em}
\end{figure*}

In this paper, we propose a method to remove waterdrops from multi-view images. When taking images with waterdrop adhered on the glass, we observe two typical scenarios: the waterdrops remain fixed in the scene while the camera is moving, and the waterdrops are static relative to the camera. The former case is often encountered when users take images through a window covered by waterdrops, while the latter is common for the cameras installed on the autonomous vehicles where the waterdrops may fall on the lens. Our method is able to effectively handle both cases. Fig. \ref{fig2} presents the overview of our method. It first detects the image regions covered by waterdrops from a deep waterdrop detector, and then excludes those regions from the training of NeRF. We will detail each component as follows.

\subsection{Background on NeRF} 
Given a set of input multi-view images (together with both the camera intrinsic and extrinsic parameters), NeRF \cite{mildenhall2021nerf} first transfers the pixels in the input images into rays using the estimated camera poses from structure from motion (SfM)\cite{schonberger2016structure}\cite{sitzmann2019scene}. It then samples points along each ray, and takes 5D vectors (i.e., the 3D position of sampled point and the 2D viewing directions) as input. The volume density $\sigma$ and view-dependent RGB color $\textbf{c}$ are then estimated by a Multi-layer Perceptron (MLP). The reason that NeRF predicts color from both position and viewing direction is to better deal with the specular reflection of the scene. After obtaining the volume density and color of each sampled point along the ray, it employs a conventional volumetric rendering technique to integrate the density and color to synthesize the corresponding pixel intensity $\hat{C}$ of the image. The whole process can be formally defined via following equation:
\begin{equation}
    \hat{C}(\textbf{r})=\int_{t_n}^{t_f}T(t)\sigma(\textbf{r}(t))\textbf{c}(\textbf{r}(t),\textbf{d})dt,
\end{equation}
where $t_n$ and $t_f$ are near and far bounds in volumetric rendering respectively, $\textbf{r}(t)$ is the sampled 3D point along the ray $\textbf{r}$ at the distance $t$ from the camera center,  $\sigma(\textbf{r}(t))$ represents the predicted density of the sampled point $\textbf{r}(t)$ by the MLP, $T(t)$ denotes the accumulated transmittance along the ray from $t_n$ to $t$, and is defined as $\exp(-\int_{t_n}^{t}\sigma(\textbf{r}(s))ds)$, $\textbf{d}$ is the viewing direction in the world coordinate frame, and $\textbf{c}(\textbf{r}(t),\textbf{d})$ is the predicted color of the sampled point $\textbf{r}(t)$ by the MLP.

The photometric loss, i.e. the mean squared error (MSE) between the rendered pixel intensity and the real captured intensity, is usually used to train the networks:
\begin{equation}
    \mathcal{L}=\sum_{\textbf{r}\in\mathcal{R}}\left\|\hat{C}(\textbf{r})-C(\textbf{r})\right\|^{2},
\end{equation}
where both $C(\textbf{r})$ and $\hat{C}(\textbf{r})$ denote the real captured  and  rendered pixel intensities for ray \textbf{r} respectively, and $\mathcal{R}$ denotes the set of sampled rays.

\subsection{AttGAN}

For raindrop detector, we employ the AttGAN model proposed by Qian et al.\cite{qian2018attentive}. AttGAN is a GAN-based single image waterdrop removal network. The generator of AttGAN incorporates a waterdrop detection module based on long-short term memory network (LSTM)\cite{shi2015convolutional}, and a waterdrop removal module based on U-net\cite{ronneberger2015u}. While AttGAN achieves satisfactory performance in waterdrop detection, its waterdrop removal module suffers from certain deficiencies, resulting in suboptimal output images. For example, AttGAN cannot fully eliminate the distortions caused by large waterdrops, leaving the watermark-like effect. In our work, we adopt a pre-trained waterdrop detection module from the AttGAN model as our waterdrop detector.

\subsection{3D Scene Estimation from Waterdrop Images}

The proposed DerainNeRF takes multi-view waterdrop images as input, then recovers the scene without effect of waterdrops. To achieve this, we feed the input waterdrop images into the pre-trained waterdrop detector, i.e. AttGAN as mentioned in previous section. The detector returns an attention map, where the attention value $\textbf{A}(u,v) \in [0,1]$ of pixel at $(u,v)$ indicates the probability whether it is covered by the waterdrop. We then train a NeRF to estimate scenes and block the waterdrop-covered pixels using the binary masks generated from attention maps, as shown in Fig. \ref{fig2}. For each image, we generate a binary mask $\textbf{M}$ from the following equation

\begin{equation}
  \textbf{M}(u,v)=f(\textbf{A}(u,v), t)
  \label{eq:important}
\end{equation}
where $f$ is a binary thesholding function, $\textbf{A}(u,v)$ indicates the attention of the pixel locating at $(u,v)$ coordinate, $t$ is a pre-defined threshold. To further improve the quality of the generated masks, we perform an image dilation operation on the generated binary masks. Fig. 3 shows an example result.

\begin{figure}[]
  
  \begin{subfigure}{0.32\linewidth}
    \centering
    \includegraphics[width=0.99\linewidth]{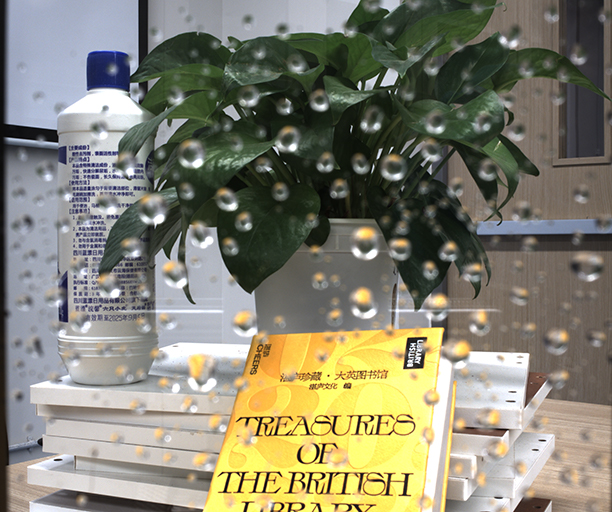}
    \caption{Input image}
    \label{fig3(a)}
  \end{subfigure}
  \begin{subfigure}{0.32\linewidth}
    \centering
    \includegraphics[width=0.99\linewidth]{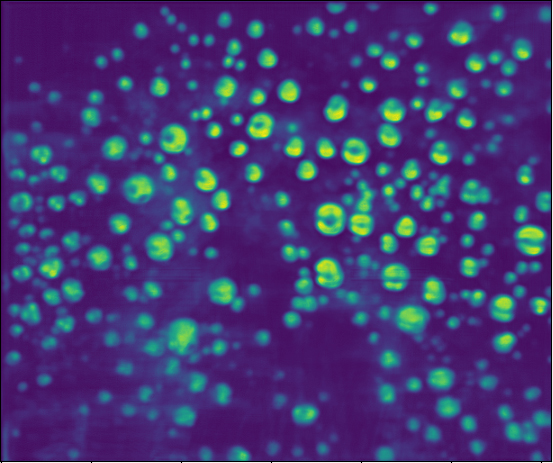}
    \caption{Attention map}
    \label{fig3(b)}
  \end{subfigure}
  \begin{subfigure}{0.32\linewidth}
    \centering
    \includegraphics[width=0.99\linewidth]{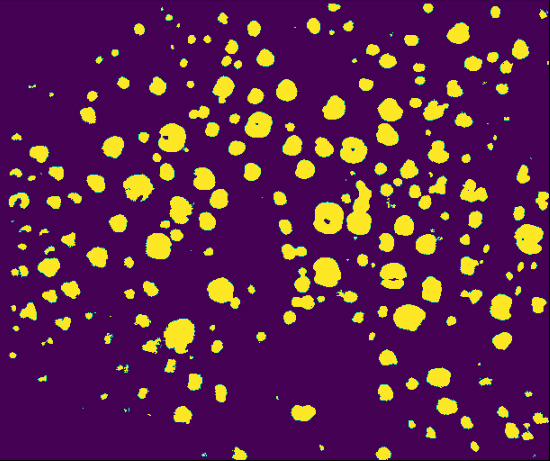}
    \caption{Mask}
    \label{fig3(c)}
  \end{subfigure}
  \caption{An example of (a) waterdrop image, (b) attention map and (c) generated binary mask from attention map}
  \label{fig3}
\end{figure}

With the generated binary masks, we train NeRF by masking the waterdrop-covered pixels in the photometric loss:

\begin{equation}
    \mathcal{L}=\sum_{\textbf{r}\in\mathcal{R}}\left\|(\hat{C}(\textbf{r})-C(\textbf{r})) \odot (\textbf{1}-\textbf{M})\right\|^{2}
\end{equation}
where $\odot$ represents element-wise multiplication, \textbf{1} is an all-one vector with the same dimension as binary mask $\textbf{M}$. For pixels covered by waterdrops, mask value would be 1, and corresponding loss would be 0, which indicates that these pixels will not contribute to the NeRF optimization. 

Furthermore, when the waterdrops are adhered on camera lens, the majority of the droplets remain stationary or move slowly within camera's field of view. Therefore we can further enhance the generated mask by averaging over multiple consecutive attention maps. This is due to our observation that certain input images might contain areas that a small portion of waterdrops cannot be detected by the network due to the overexposure or underexposure on these pixels. But in other images, such as images taken from another view, the detection returns to normal as the previously overexposed or underexposed pixels are no longer affected. Therefore, when estimating the scenes with waterdrops lying on the camera lens, we calculate an additional attention map. This attention map is the mean of all generated per-frame attention maps, and it serves to compute the binary mask:

\begin{equation}
 \textbf{M}(u,v)=f(\textbf{A}(u,v), t) \vee f( \frac{1}{n} \sum_{i=1}^{n}{\textbf{A}_i(u,v)}, t)
  \label{eq:important}
\end{equation}
where $\textbf{A}_i(u,v)$ is the attention map of $i$-th input image, and $\vee$ denotes logical OR operation. After the training process is complete, we are able to recover the clear implicit 3D scene without waterdrops. Given an arbitrary camera pose, we can render the clear novel view images following (1).

\section{EXPERIMENTS}

\subsection{Implementation Details}

To obtain the pre-trained waterdrop detector from AttGAN, we first train the AttGAN following the guidelines of Qian et al.\cite{qian2018attentive}, from the dataset in AttGAN paper. When generating the binary masks from attention maps, we set the threashold value $t$ between 0.2 and 0.4, depending on the resolution of input images. For DerainNeRF training, we use ADAM optimizer\cite{kingma2017adam} with learning rate decays from $5\times10^{-4}$ to $5\times10^{-5}$ exponentially. We train our model for 200K iterations, with 1024 rays as batch size, on an NVIDIA GeForce RTX 3090 GPU. We use COLMAP\cite{schonberger2016structure}, a popular SfM/MVS software to estimate camera poses from input images prior to the training procedure.

\subsection{Datasets}
To assess the effectiveness of DerainNeRF, we conduct evaluations on both the synthetic and real datasets. The synthetic dataset comes from the Blender scenes used in Deblur-NeRF\cite{ma2022deblur}. We select 5 virtual scenes, then add the physically simulated waterdrops to the scene via Blender and capture the multi-view images from different camera poses. We capture images under two scenarios: (a) scenes with waterdrops fixed to the scene while the camera is moving (denoted as ``-move" in the dataset), and (b) scenes with waterdrops fixed to the camera lens.

For the collection of real datasets, we setup a hardware system. We use a HIKROBOT MV-CA050-12UC camera to capture images. During image acquisition process, we place a 3mm thick glass in front of the camera lenses and spray waterdrops on the glass. We simulate both types of scenarios with this setup (i.e., to move camera only or move camera and glass simultaneously). We have also tested the proposed method with outdoor real dataset, where images are taken from a moving vehicle under rainy weather conditions. In the outdoor dataset, the waterdrop-covered glass is fixed to the camera. Each scene in the synthetic and real dataset comprises 20-25 images.

\subsection{Results}
\begin{figure*}[ht]

\hspace{-0.25cm}
\begin{minipage}[c]{1.0\textwidth}
\begin{minipage}[c]{\linewidth}
\centering
  \begin{minipage}[c]{0.123\linewidth}
  \centering
  \small
  \ \\ \ Input
  \end{minipage}
  ~
  \begin{minipage}[c]{0.123\linewidth}
  \centering
  \small
  \ \\ NeRF~\cite{mildenhall2021nerf}
  \end{minipage}
  ~
  \begin{minipage}[c]{0.124\linewidth}
  \centering
  \small
  \ \\ \ AttGAN~\cite{qian2018attentive}
  \end{minipage}
  ~
  \begin{minipage}[c]{0.123\linewidth}
  \centering
  \small
  \ \\ \ Quan et al.~\cite{quan2019deep}
  \end{minipage}
  ~
  \begin{minipage}[c]{0.123\linewidth}
  \centering
  \small
  \ \\ \ Wen et al.~\cite{wen2023video}
  \end{minipage}
  ~
  \begin{minipage}[c]{0.123\linewidth}
  \centering
  \small
  \ \\ \ Ours
  \end{minipage}
  ~
  \begin{minipage}[c]{0.123\linewidth}
  \centering
  \small
  \ \\ \ Ground Truth
  \end{minipage}
\end{minipage}
\\
\begin{minipage}[c]{\linewidth}
\centering
  \begin{minipage}[c]{0.123\linewidth}
  \includegraphics[width=1.1\linewidth]{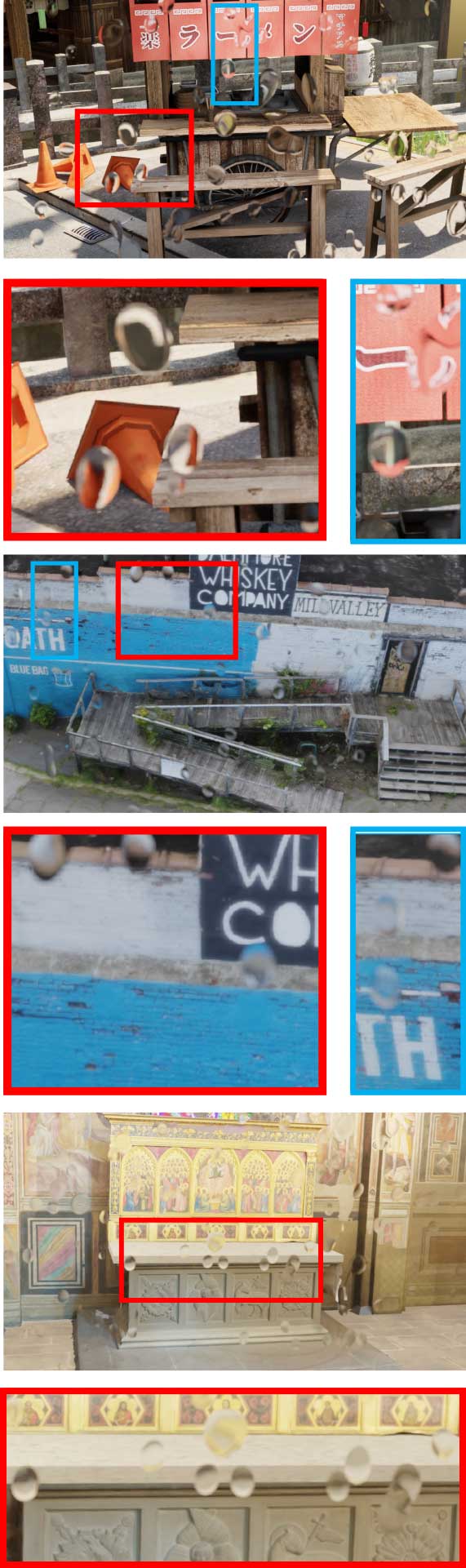}
  
  \end{minipage}
  ~
  \begin{minipage}[c]{0.123\linewidth}
  \includegraphics[width=1.1\linewidth]{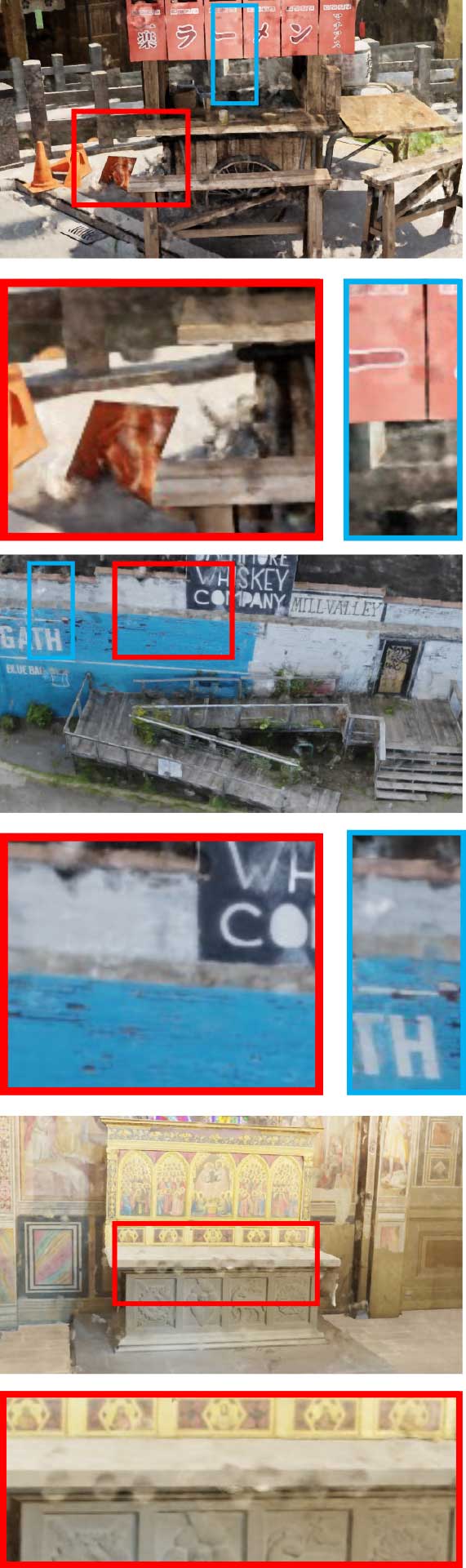}
  \end{minipage}
  ~
  \begin{minipage}[c]{0.124\linewidth}
  \includegraphics[width=1.1\linewidth]{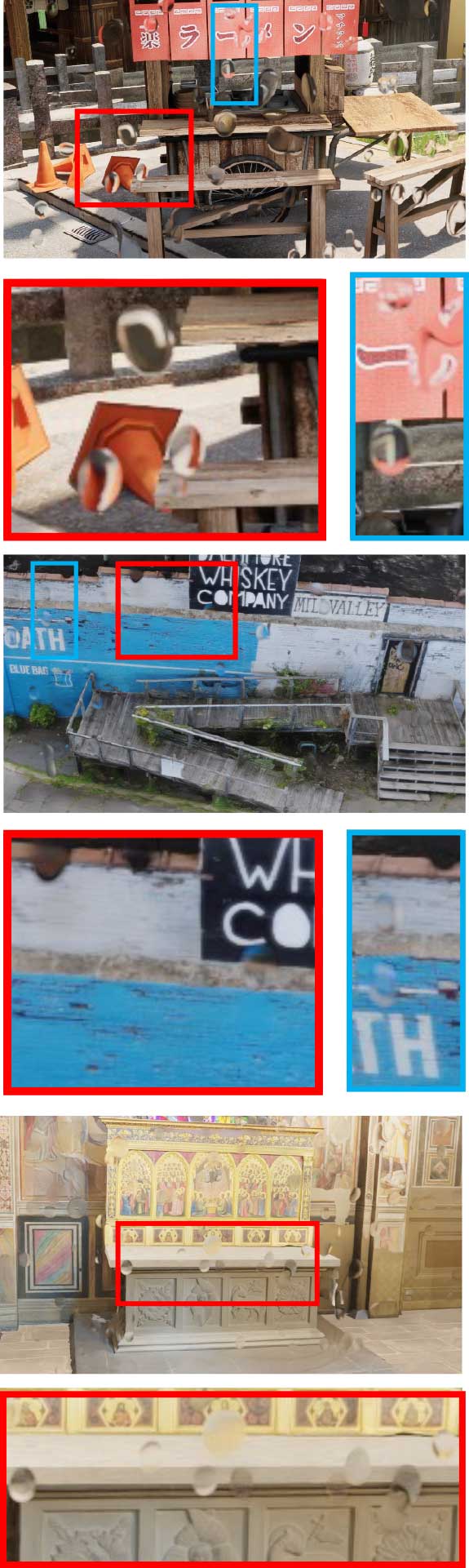}
  \end{minipage}
  ~
  \begin{minipage}[c]{0.123\linewidth}
  \includegraphics[width=1.1\linewidth]{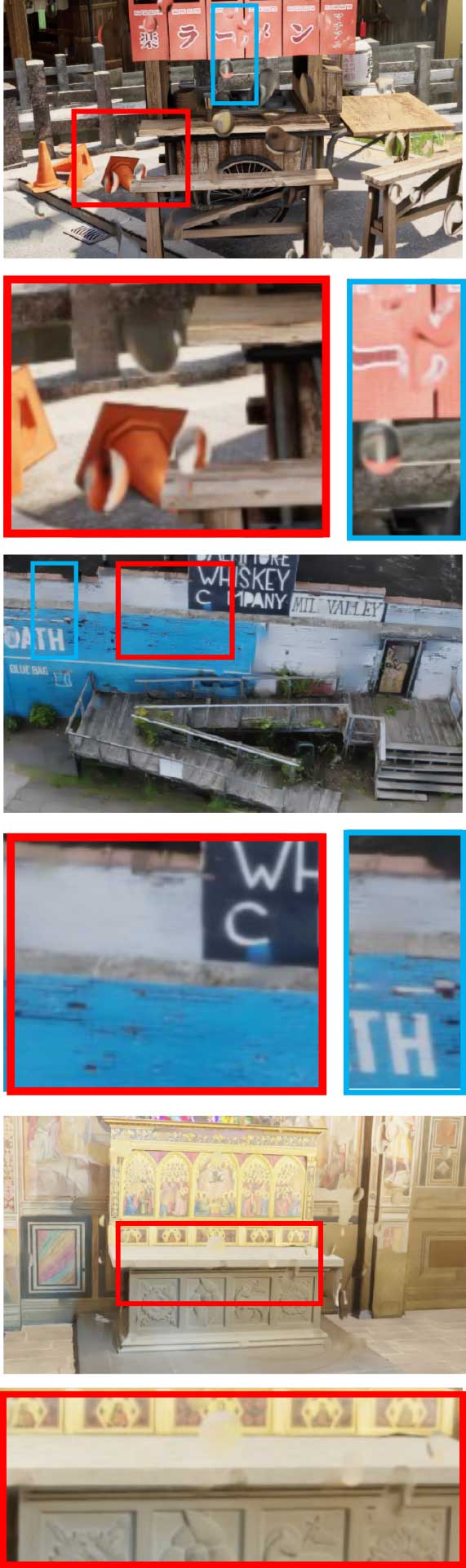}
  \end{minipage}
  ~
  \begin{minipage}[c]{0.123\linewidth}
  \includegraphics[width=1.1\linewidth]{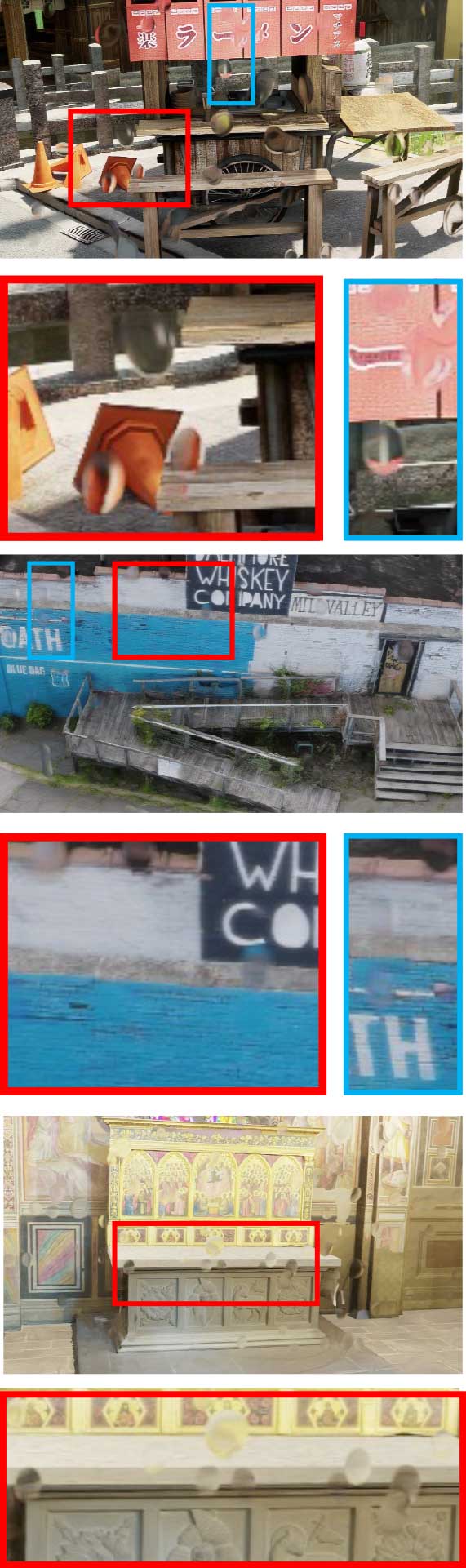}
  \end{minipage}
  ~
  \begin{minipage}[c]{0.123\linewidth}
  \includegraphics[width=1.1\linewidth]{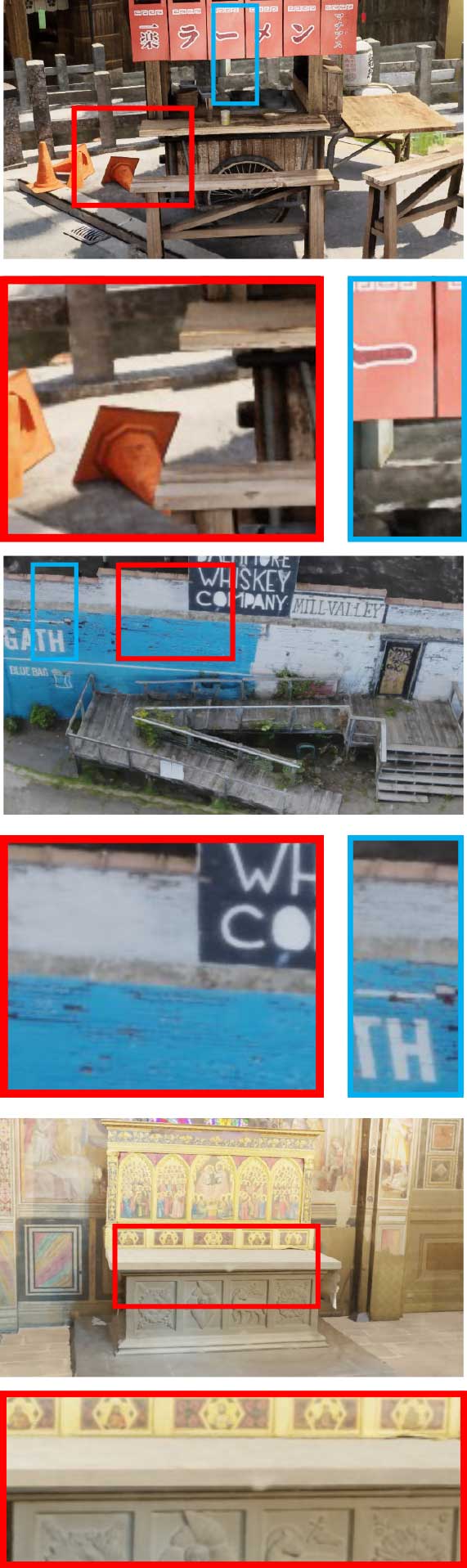}
  \end{minipage}
  ~
  \begin{minipage}[c]{0.123\linewidth}
  \includegraphics[width=1.1\linewidth]{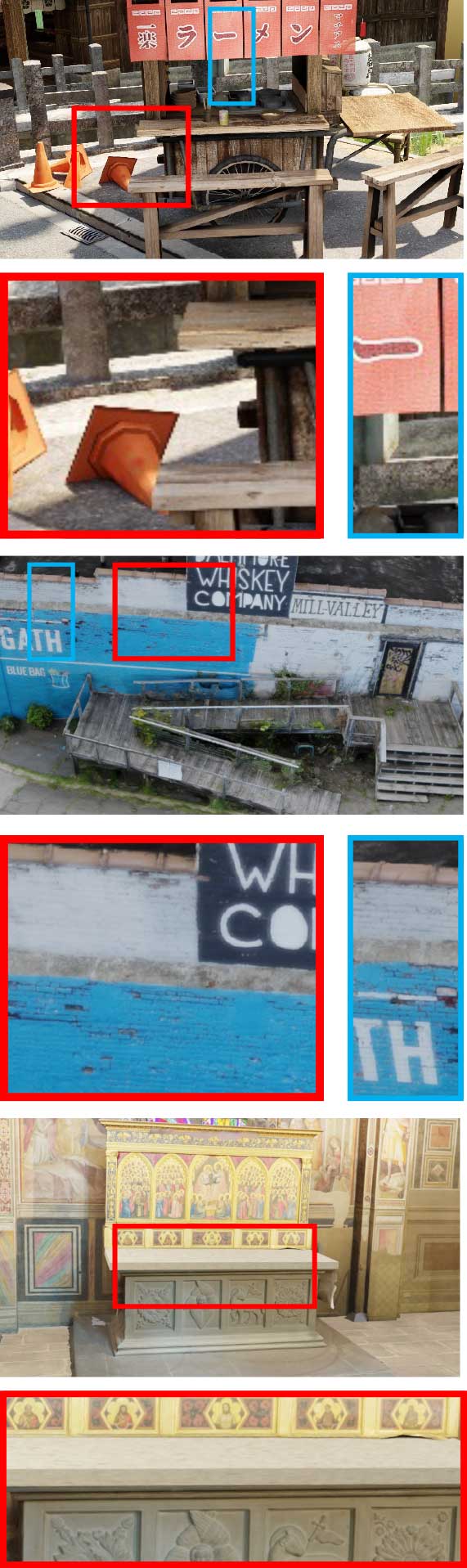}
  \end{minipage}
\end{minipage}
\end{minipage}
\\

   \caption{Qualitative evaluations of our method against SOTA image waterdrop removal methods on the synthetic dataset. Top to bottom shows different scenes including \textit{Tanabata}, \textit{Factory} and \textit{Church}. We render waterdrop-removed images from clear 3D scenes estimated by our method.
   }
   \label{fig4}
\end{figure*}
\begin{figure*}[ht]
\hspace{-0.3cm}
\begin{minipage}[c]{1.01\textwidth}
\begin{minipage}[c]{\linewidth}
\centering
  \begin{minipage}[c]{0.145\linewidth}
  \centering
  \small
  \ Input
  \end{minipage}
  ~
  \begin{minipage}[c]{0.145\linewidth}
  \centering
  \small
  \ NeRF~\cite{mildenhall2021nerf}
  \end{minipage}
  ~
  \begin{minipage}[c]{0.145\linewidth}
  \centering
  \small
  \ AttGAN~\cite{qian2018attentive}
  \end{minipage}
  ~
  \begin{minipage}[c]{0.145\linewidth}
  \centering
  \small
  \ Quan et al.~\cite{quan2019deep}
  \end{minipage}
  ~
  \begin{minipage}[c]{0.145\linewidth}
  \centering
  \small
  \ Wen et al.~\cite{wen2023video}
  \end{minipage}
  ~
  \begin{minipage}[c]{0.145\linewidth}
  \centering
  \small
  \ Ours
  \end{minipage}
\end{minipage}
\\
\begin{minipage}[c]{\linewidth}
\centering
  \begin{minipage}[c]{0.145\linewidth}
  \includegraphics[width=1.08\linewidth]{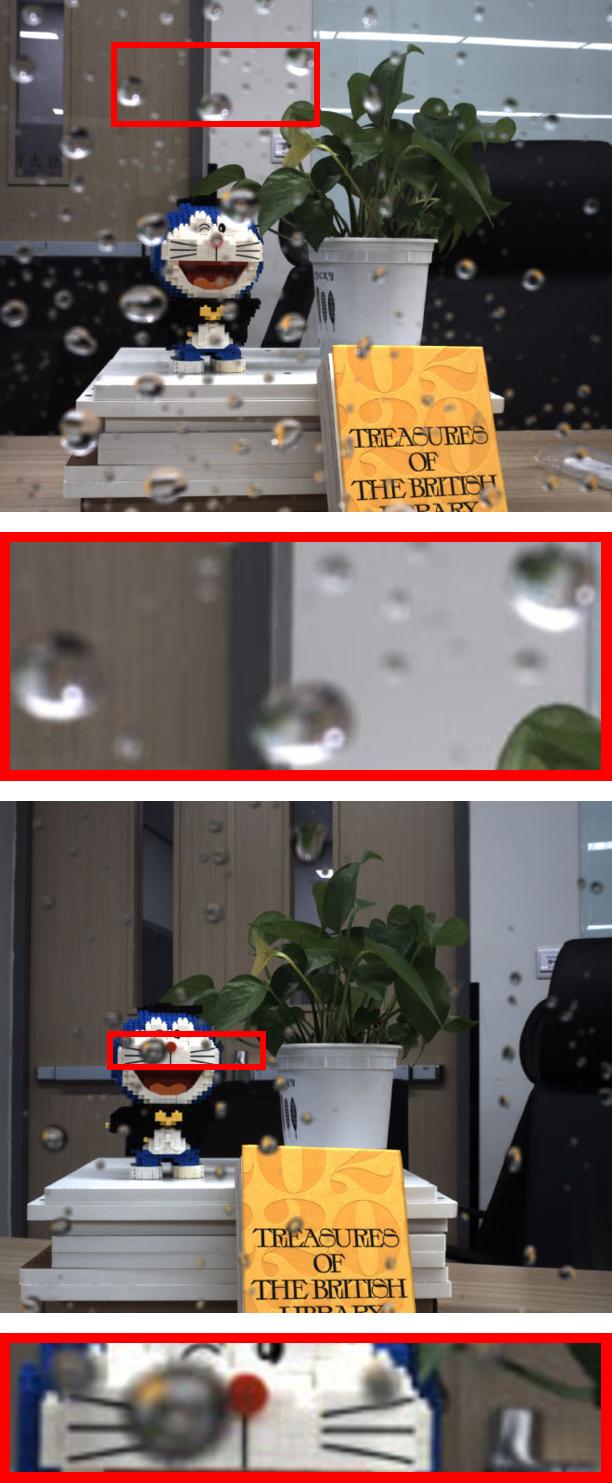}
  \end{minipage}
  ~
  \begin{minipage}[c]{0.145\linewidth}
  \includegraphics[width=1.08\linewidth]{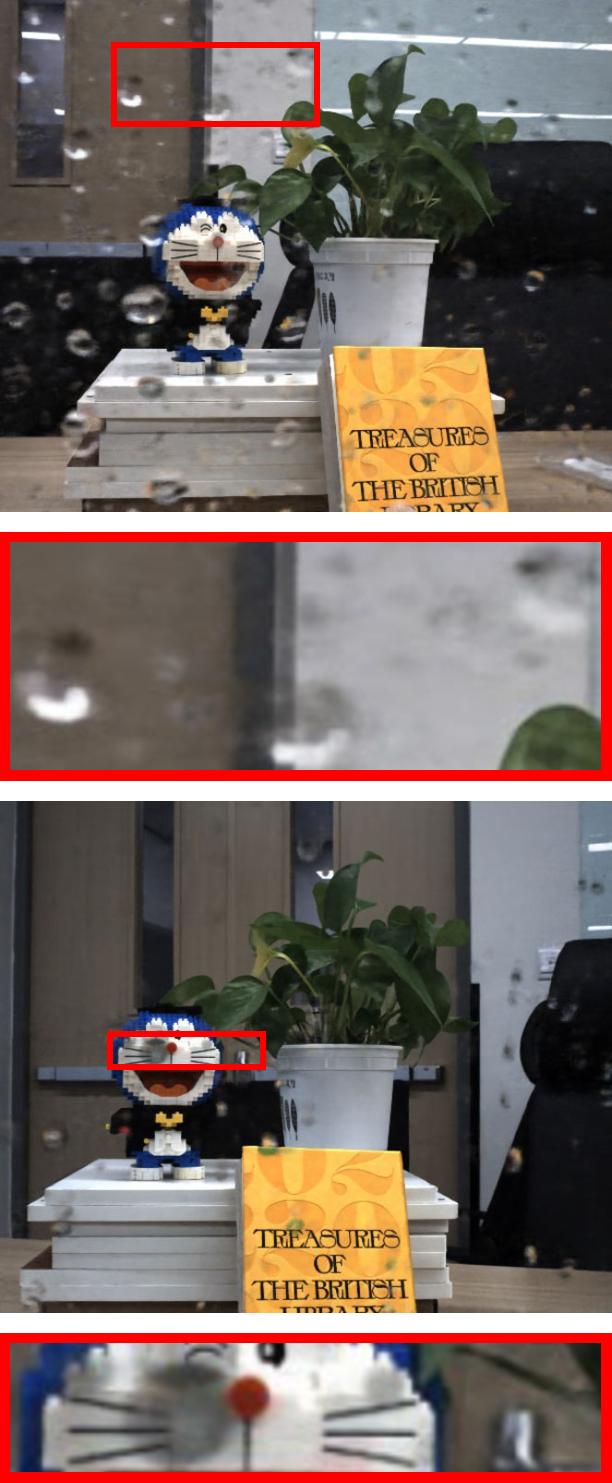}
  \end{minipage}
  ~
  \begin{minipage}[c]{0.145\linewidth}
  \includegraphics[width=1.08\linewidth]{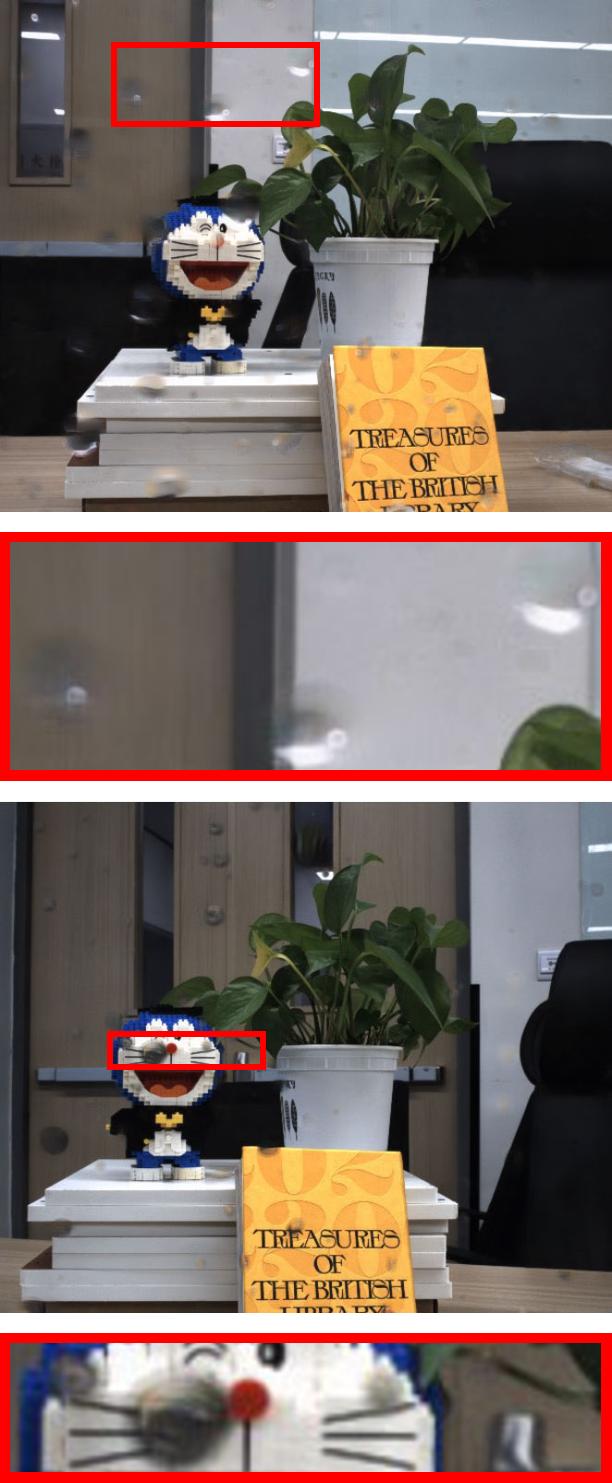}
  \end{minipage}
  ~
  \begin{minipage}[c]{0.145\linewidth}
  \includegraphics[width=1.08\linewidth]{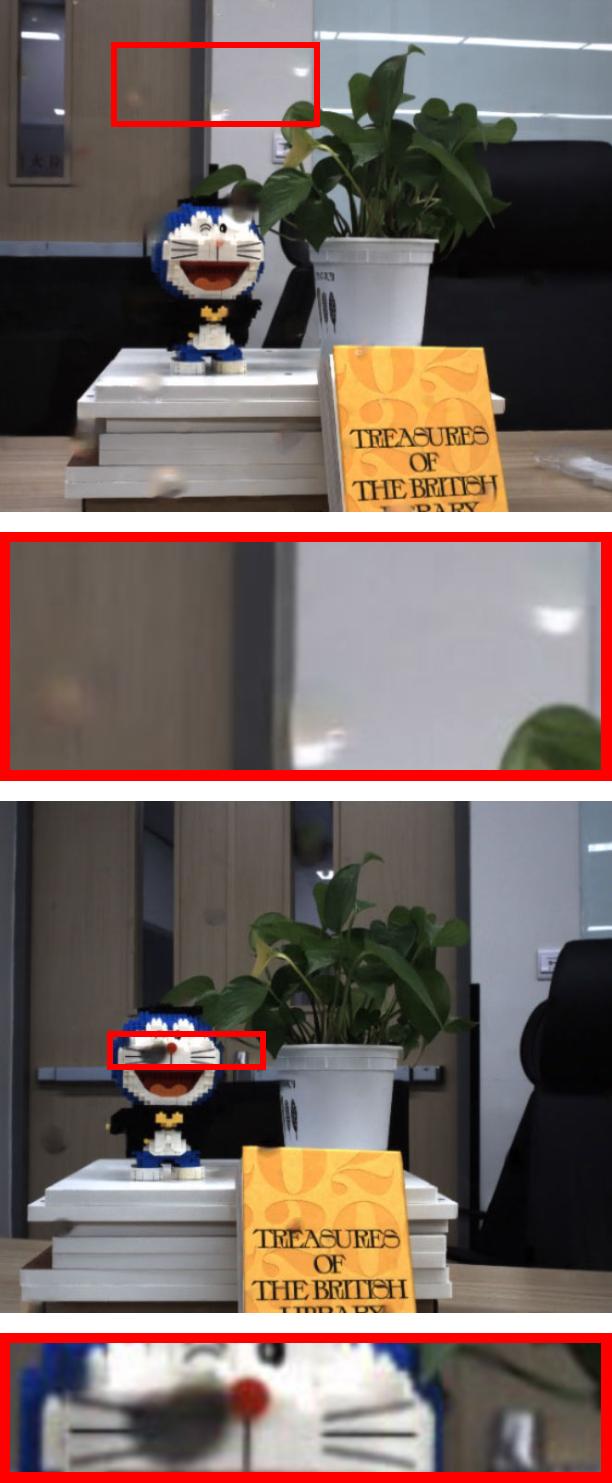}
  \end{minipage}
  ~
  \begin{minipage}[c]{0.145\linewidth}
  \includegraphics[width=1.08\linewidth]{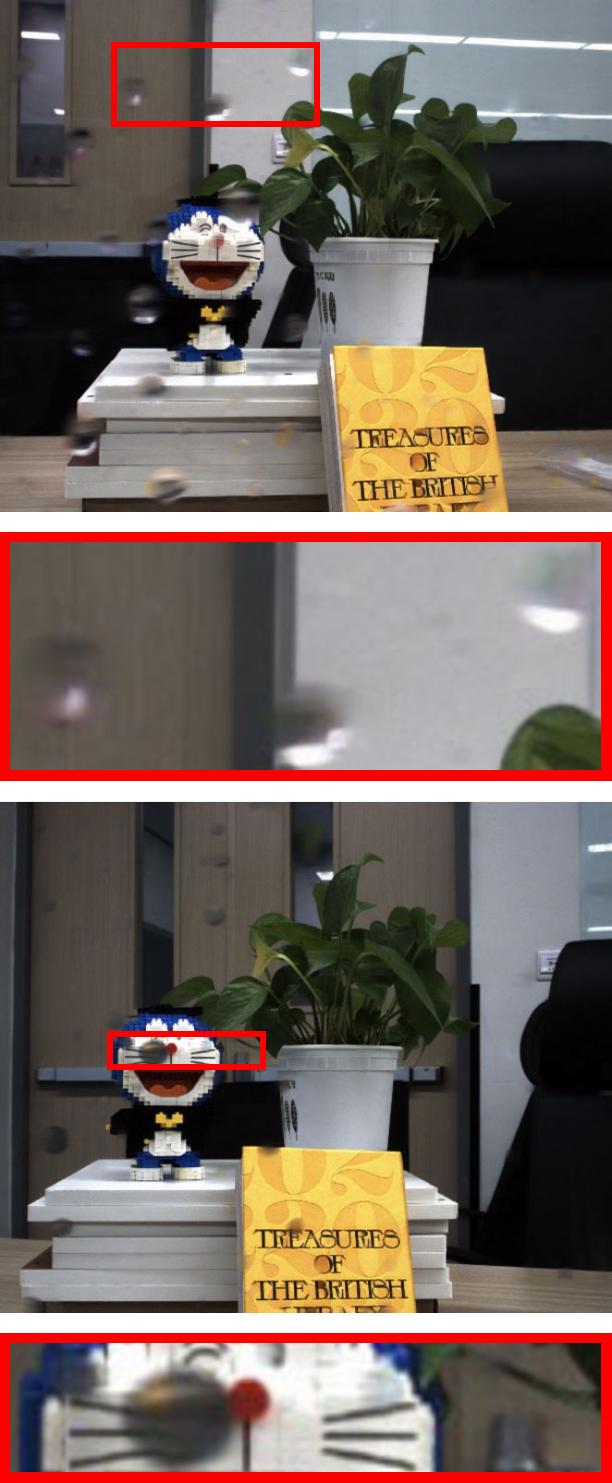}
  \end{minipage}
  ~
  \begin{minipage}[c]{0.145\linewidth}
  \includegraphics[width=1.08\linewidth]{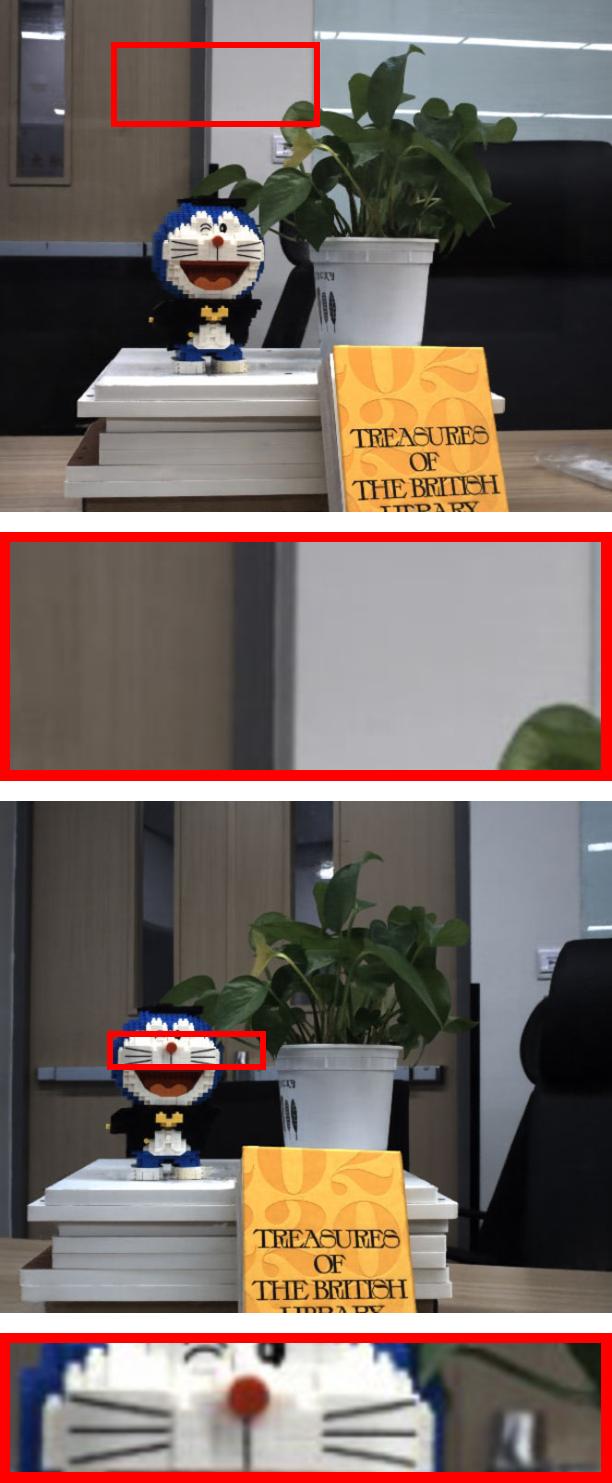}
  \end{minipage}
\end{minipage}
\end{minipage}
\\
   \caption{Qualitative comparisons between different methods with real indoor dataset. The experimental results demonstrate that our method can effectively remove droplets whether the glass with waterdrops is fixed to the scene (top row) or fixed to the camera (bottom row).
   }
   \label{fig5}
\end{figure*}
\begin{figure*}[ht]
\hspace{-0.3cm}
\begin{minipage}[c]{1.01\textwidth}
\begin{minipage}[c]{\linewidth}
\centering
  \begin{minipage}[c]{0.145\linewidth}
  \centering
  \small
  \ \\ Input
  \end{minipage}
  ~
  \begin{minipage}[c]{0.145\linewidth}
  \centering
  \small
  \ \\ NeRF~\cite{mildenhall2021nerf}
  \end{minipage}
  ~
  \begin{minipage}[c]{0.145\linewidth}
  \centering
  \small
  \ \\ AttGAN~\cite{qian2018attentive}
  \end{minipage}
  ~
  \begin{minipage}[c]{0.145\linewidth}
  \centering
  \small
  \ \\ Quan et al.~\cite{quan2019deep}
  \end{minipage}
  ~
  \begin{minipage}[c]{0.145\linewidth}
  \centering
  \small
  \ \\ Wen et al.~\cite{wen2023video}
  \end{minipage}
  ~
  \begin{minipage}[c]{0.145\linewidth}
  \centering
  \small
  \ \\ Ours
  \end{minipage}
\end{minipage}
\\
\begin{minipage}[c]{\linewidth}
\centering
  \begin{minipage}[c]{0.145\linewidth}
  \includegraphics[width=1.08\linewidth]{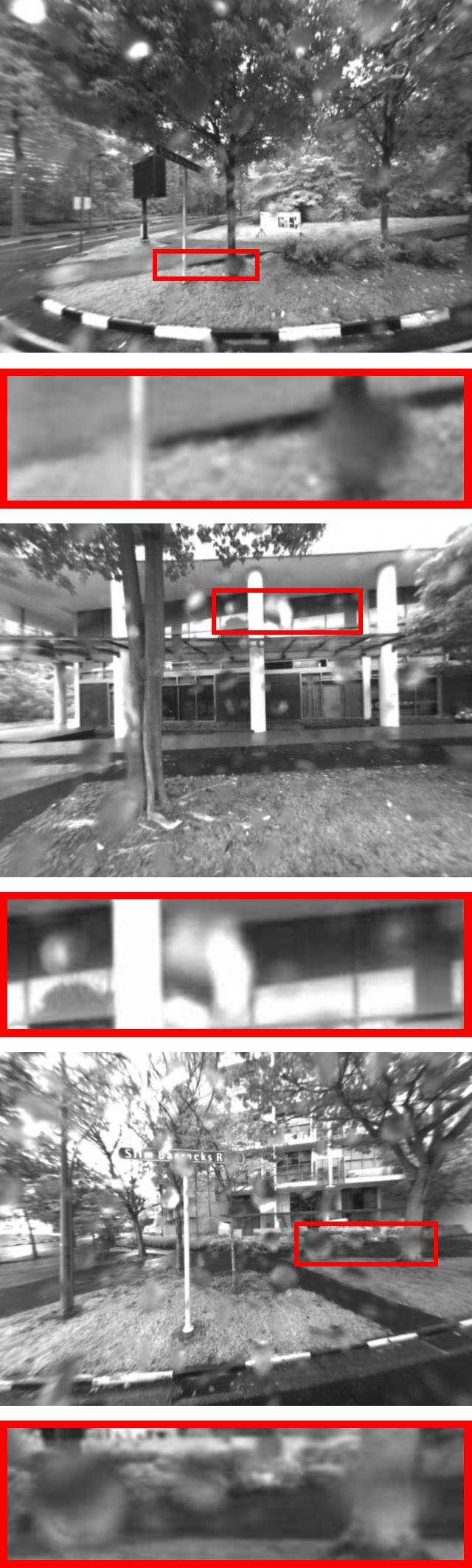}
  \end{minipage}
  ~
  \begin{minipage}[c]{0.145\linewidth}
  \includegraphics[width=1.08\linewidth]{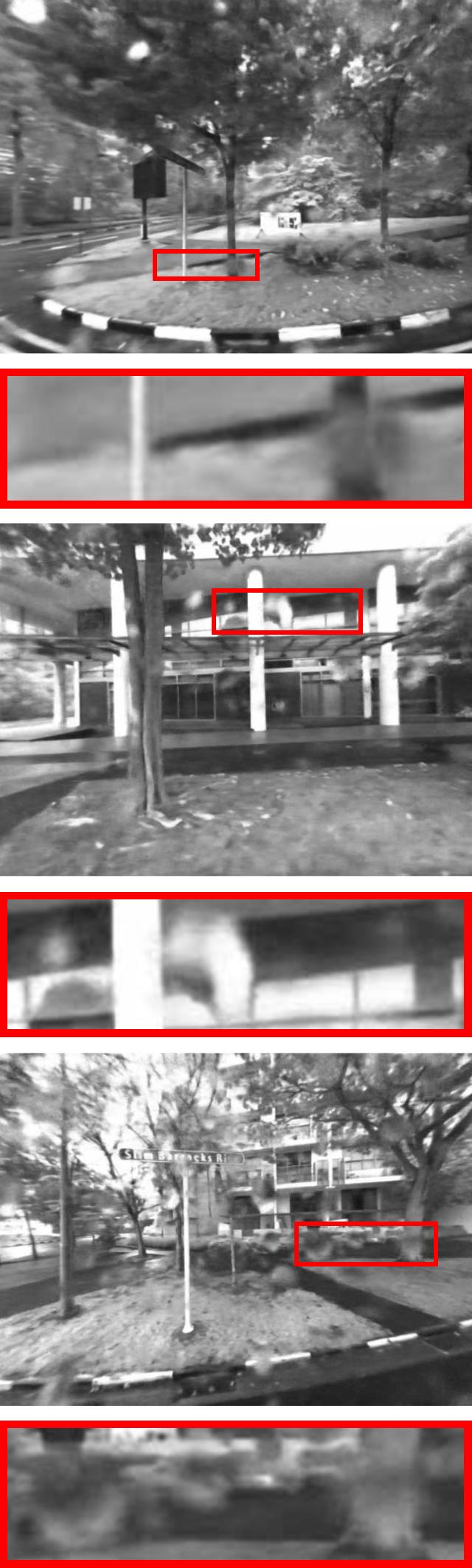}
  \end{minipage}
  ~
  \begin{minipage}[c]{0.145\linewidth}
  \includegraphics[width=1.08\linewidth]{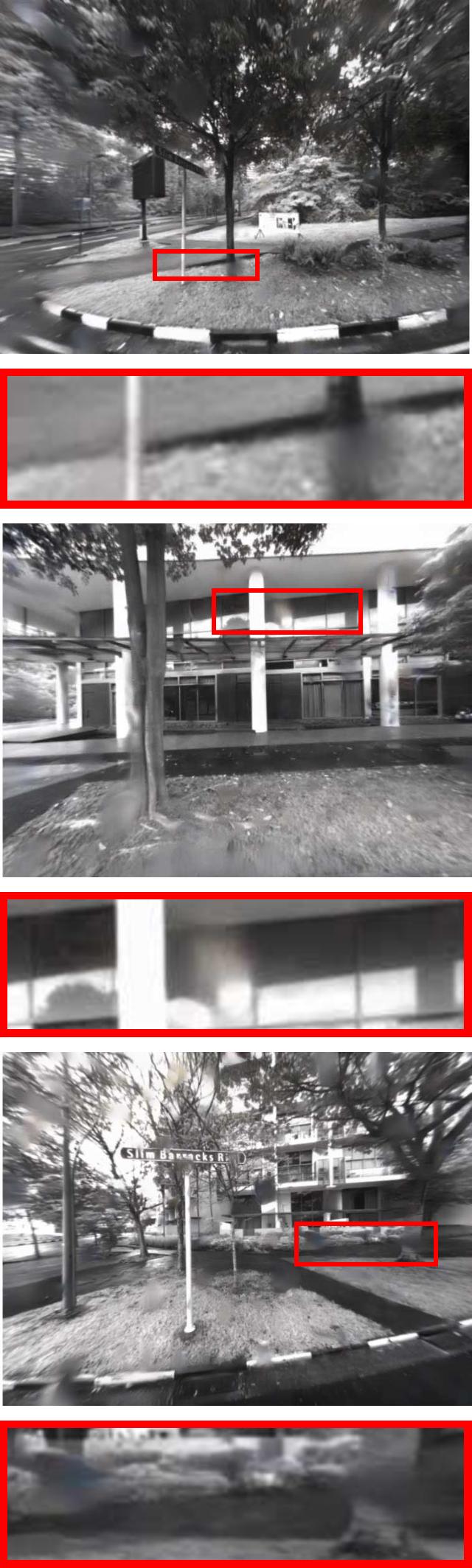}
  \end{minipage}
  ~
  \begin{minipage}[c]{0.145\linewidth}
  \includegraphics[width=1.08\linewidth]{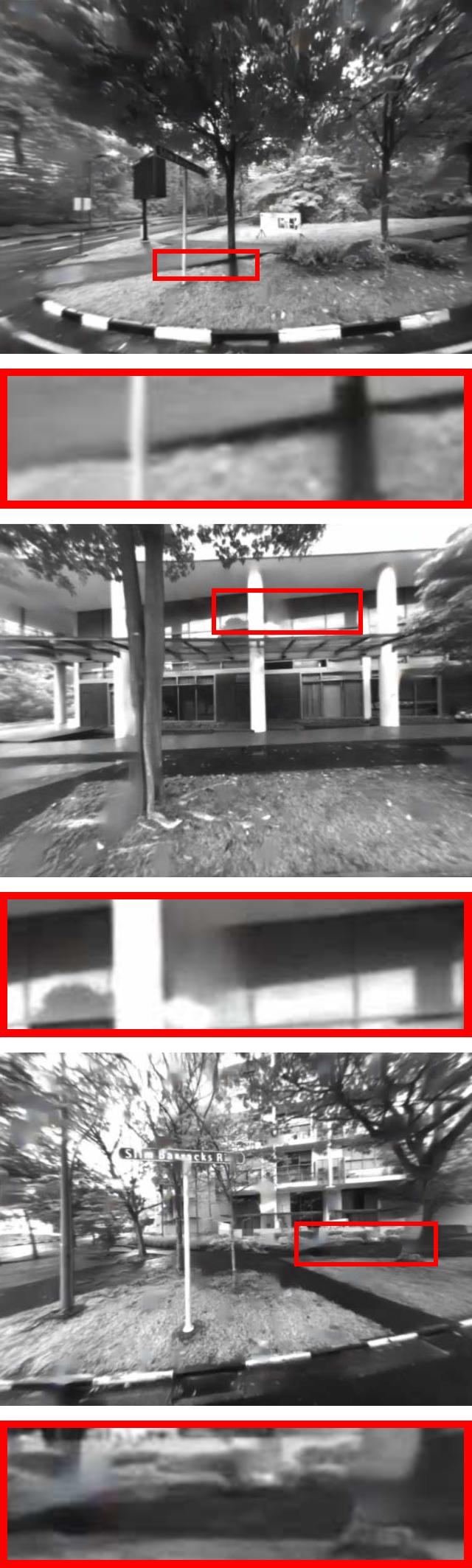}
  \end{minipage}
  ~
  \begin{minipage}[c]{0.145\linewidth}
  \includegraphics[width=1.08\linewidth]{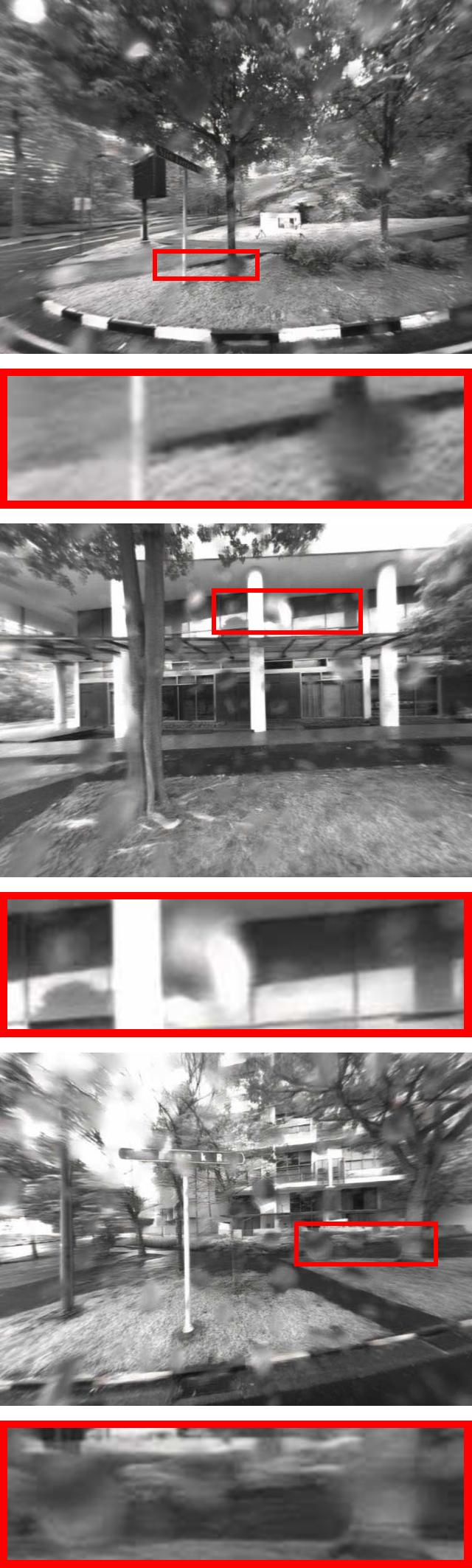}
  \end{minipage}
  ~
  \begin{minipage}[c]{0.145\linewidth}
  \includegraphics[width=1.08\linewidth]{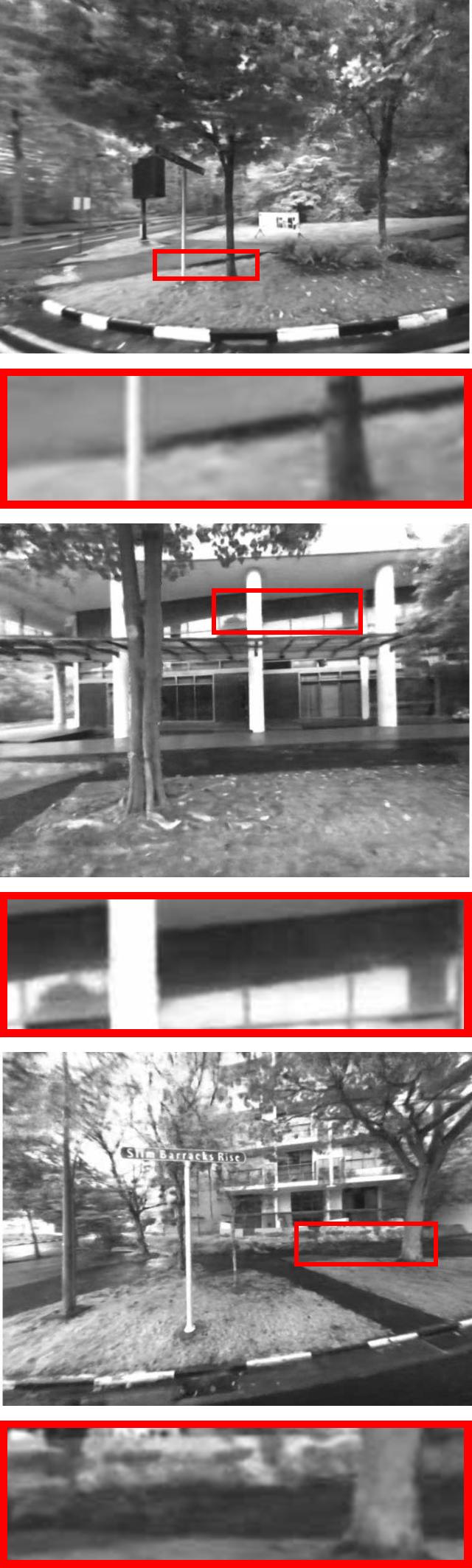}
  \end{minipage}
\end{minipage}
\end{minipage}
\\
   \caption{Qualitative comparisons between different methods on outdoor real  dataset. The experimental results demonstrate
that our method still presents a satisfying waterdrop removal performance on real outdoor dataset.
   }
   \label{fig6}
   \vspace{-0.7em}
\end{figure*}

We evaluate the proposed DerainNeRF on both synthetic and real datasets. Since DerainNeRF can render waterdrop-removed clear images from the reconstructed scene thanks to NeRF's powerful image synthesis capabilities, we compare the performance of DerainNeRF against that of vanilla NeRF, and state-of-the-art (SOTA) image waterdrop remvoval methods, including AttGAN\cite{qian2018attentive}, Quan et al.\cite{quan2019deep}, and Wen et al.\cite{wen2023video}. We also compare the performance against that of vanilla NeRF with waterdrop-removed images from prior mentioned methods. We evaluate the performance quantitatively with the commonly used metrics, such as the structural similarity index (SSIM), peak signal to noise ratio (PSNR), and learned perceptual image patch similarity (LPIPS)\cite{zhang2018unreasonable}. 

\begin{table*}
        \caption{Quantitative comparisons on the synthetic dataset. The experimental results demonstrate that our method can render clear images with higher quality than those from existing state-of-the-art image waterdrop removal methods.}
	\setlength\tabcolsep{2pt}
	\parbox{\textwidth}{
		\resizebox{\linewidth}{!}{
		\begin{tabular}{c|ccc|ccc|ccc|ccc|ccc}
			
			\hline
			& \multicolumn{3}{c|}{Tanabata} & \multicolumn{3}{c|}{Tanabata-move} & \multicolumn{3}{c|}{Factory} & \multicolumn{3}{c|}{Church} & \multicolumn{3}{c}{Church-move} \\
			& PSNR$\uparrow$ & SSIM$\uparrow$ & LPIPS$\downarrow$ & PSNR$\uparrow$ & SSIM$\uparrow$ & LPIPS$\downarrow$ & PSNR$\uparrow$ & SSIM$\uparrow$ & LPIPS$\downarrow$ & PSNR$\uparrow$ & SSIM$\uparrow$ & LPIPS$\downarrow$ & PSNR$\uparrow$ & SSIM$\uparrow$ & LPIPS$\downarrow$ \\
			\hline
			
			NeRF\cite{mildenhall2021nerf} &19.6145 &0.7700 &0.2057 &23.7986 &0.8604 &0.1097 &26.2478 &0.8715 &0.1093 &27.5391 &0.945 &0.0921 &27.3940 &0.9520 &0.0817\\
			AttGAN\cite{qian2018attentive} &19.7771 &0.8506 &0.1359 &23.7845 &{\textbf{0.9118}} &0.0775 &27.7066 &{\textbf{0.9404}} &0.0854 & 25.9504 & 0.9514 & 0.0752 & 26.6045 & 0.9525 & 0.0760\\
			Quan et al.\cite{quan2019deep} &20.6615 & 0.8465 & 0.1359 & 23.6408 & 0.8816 & 0.1413 & 25.0735 & 0.8899 & 0.1369 & 27.3451 & 0.9500 & 0.0752 & 26.6356 & 0.9512 & 0.1007\\
			Wen et al.\cite{wen2023video} &20.9196 & 0.8399 & 0.1293  & 23.2868 & 0.8784 & 0.0923 & 27.5619 & 0.9153 & 0.0815 & 27.7761 & 0.9462 & 0.0750 & 27.6249 & 0.9536 & 0.0726\\
            NeRF+AttGAN &20.5404 & 0.7938 & 0.1923  & 23.0334 & 0.8566 & 0.0975 & 26.1938 & 0.8804 & 0.0864 & 26.2998 & 0.9432 & 0.0764 & 27.2996 & 0.9523 & 0.0596\\
            NeRF+Quan et al. &20.9633 & 0.7916 & 0.2456  & 23.4550 & 0.8394 & 0.1823 & 26.4362 & 0.8646 & 0.1569 & 27.3061 & 0.9369 & 0.1231 & 27.3506 & 0.9471 & 0.1048\\
            NeRF+Wen et al. &21.2252 & 0.7912 & 0.1796  & 23.1937 & 0.8371 & 0.1193 & 27.7594 & 0.8739 & 0.1548 & 28.0618 & 0.9445 & 0.0771 & 27.3503 & 0.9518 & 0.0776\\
			\hline
			
			DerainNeRF (ours) &{\textbf{26.0367}} &{\textbf{0.8866}} &{\textbf{0.1081}} &{\textbf{24.5683}} &0.8882 &{\textbf{0.0613}} &{\textbf{30.8504}} & 0.9229 &{\textbf{0.0564}} &{\textbf{30.1028}} &{\textbf{0.9676}} &{\textbf{0.0426}} &{\textbf{28.1776}} &{\textbf{0.9570}} &{\textbf{0.0562}}\\
            \hline
			
	\end{tabular}}

	\label{table I}}

	\vspace{-1.7em}
\end{table*}

The experimental results on the synthetic dataset provides empirical evidence of the efficacy of DerainNeRF in eliminating waterdrops and reconstructing visually clear 3D scenes with high-fidelity images, as shown in both Fig. \ref{fig4} and Table I. It is noteworthy that, in certain scenes, the structural similarity index (SSIM) of our method's images does not exceed that of AttGAN. This observation can be attributed to the fact that our approach does not directly generate waterdrop-free images from the input; instead, it utilizes NeRF to render clear images based on the underlying scene representation. While NeRF successfully preserves the majority of scene details, the rendering process may introduce a marginal loss of image information. Nevertheless, our method exhibits significantly superior performance in terms of peak signal-to-noise ratio (PSNR) and learned perceptual image patch similarity (LPIPS). 

To evaluate the performance of DerainNeRF on real datasets, we also conduct qualitative comparisons against state-of-the-art methods. Fig. \ref{fig5} and Fig. \ref{fig6} illustrate the comparisons between methods, depicting the outcomes for real indoor and outdoor datasets, respectively. Notably, existing state-of-the-art techniques exhibit limitations when confronted with large waterdrops, resulting in noticeable deficiencies in the output images. In contrast, our DerainNeRF surpasses these methods on real datasets by effectively removing waterdrops in various sizes and shapes.

 \subsection{Ablation Study}
 
\begin{figure}[]
  
   \begin{subfigure}{0.49\linewidth}
     \centering
     \includegraphics[width=0.99\linewidth]{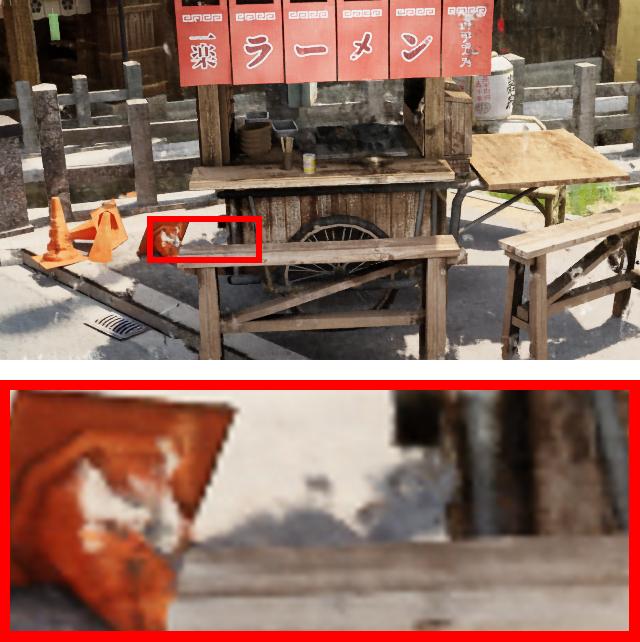}
     \caption{Ours (no enhancement)}
     \label{fig7(a)}
   \end{subfigure}
   \begin{subfigure}{0.49\linewidth}
     \centering
     \includegraphics[width=0.99\linewidth]{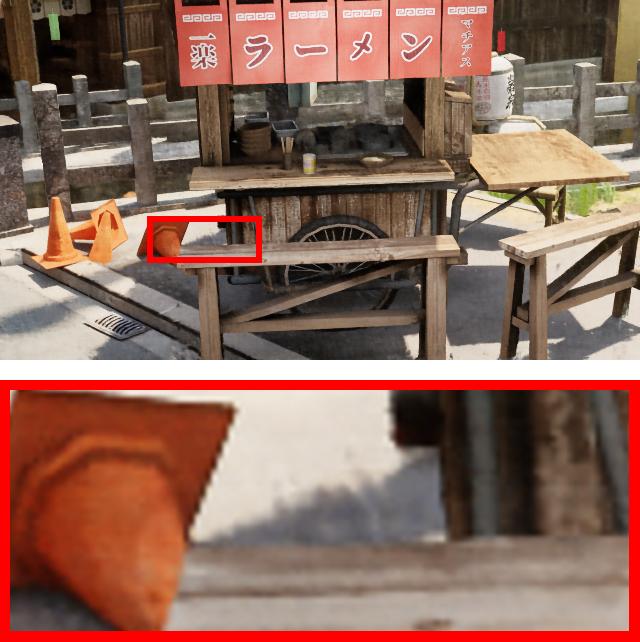}
     \caption{Ours (full)}
     \label{fig7(b)}
   \end{subfigure}
   \caption{Qualitative comparison on synthetic dataset for our ablation study.}
   \label{fig7}
   \vspace{-0.7em}
 \end{figure}

 To better analyze the effectiveness of mask enhancement through average attention map, we compare the results with and without mask enhancement procedure described in (5). We conduct comparisons on synthetic \textit{Tanabata, Factory} and \textit{Church} dataset, where the waterdrops are static relative to the camera. Fig. \ref{fig7} shows the qualitative comparisons between (a) DerainNeRF without mask enhancement and (b) full pipeline. Table II presents quantitative comparisons, where the structural similarity index (SSIM), peak signal to noise ratio (PSNR) and learned perceptual image patch similarity (LPIPS) are the mean value of those on three synthetic datsets. As is shown in Fig. \ref{fig7} and Table II, the quality of synthesized images from our full pipeline is significantly better than the images rendered from the scene represented by DerainNeRF without mask enhancement. Therefore, mask enhancement based on the average attention map has positive effect during the training of DerainNeRF, especially when the waterdrops become dense.  

\begin{table}[]
     \caption{Ablation study on synthetic \textit{Tanabata, Factory} and \textit{Church} dataset.}
     \resizebox{1.0\linewidth}{!}{
    \footnotesize
    \begin{tabular}{c|ccc}
     \hline
     & \multicolumn{3}{c}{\scriptsize{ Tanabata, Factory and Church}} \\
    & \scriptsize PSNR$\uparrow$ &\scriptsize SSIM$\uparrow$ & \scriptsize LPIPS$\downarrow$\\
 	\hline
    \scriptsize Ours (no enhancement) & \scriptsize 27.2611 & \scriptsize 0.9131 & \scriptsize 0.0922 \\
 	 \scriptsize Ours (full) & \scriptsize {\textbf{28.3549}} & \scriptsize {\textbf{0.9299}} & \scriptsize {\textbf{0.0687}} \\
			
     \hline
    \end{tabular}
     }
     \label{tab:my_label}
     \vspace{-2em}
\end{table}

\section{CONCLUSIONS}

In this paper, we introduce DerainNeRF, a novel approach for 3D scene estimation from multi-view waterdrop degraded images with NeRF representation. DerainNeRF addresses the challenge of waterdrop removal by utilizing a comprehensive pipeline. Initially, a pre-trained waterdrop detector is employed to identify and localize waterdrops within the input images. Subsequently, our approach estimates clear scenes by leveraging a NeRF-based network exploiting non-occluded pixels. To validate the effectiveness of our proposed method, we conduct a thorough evaluation against existing state-of-the-art techniques for image waterdrop removal with both synthetic and real datasets. The experimental results demonstrate the superior performance of our method in comparison to existing approaches.





{\small
\bibliographystyle{IEEEtran}
\bibliography{root}
}

\end{document}